\documentclass[aps,prl,floatfix,twocolumn,showpacs,10pt]{revtex4-2}

\usepackage{graphicx}
\usepackage{dcolumn}
\usepackage{bm}
\usepackage{amssymb}
\usepackage{amsmath} 
\usepackage{rotating} 
\usepackage{xcolor}
\usepackage{hyperref}
\usepackage[mathlines]{lineno}
\usepackage{booktabs}

\begin{document}

\preprint{draft V4}

\title{Self-consistent Validation for Machine Learning Electronic Structure}


\author{Gengyuan Hu}
 \thanks{Main contribution}
 \email{hugengyuan@pjlab.org.cn}
\author{Gengchen Wei}
\author{Zekun Lou}
\author{Philip H.S. Torr} 
\author{Wanli Ouyang}
\author{Han-sen Zhong}
\author{Chen Lin}
 \thanks{Main contribution}
 \email{chen.lin@eng.ox.ac.uk}
\affiliation{Shanghai Artificial Intelligence Laboratory}%
\affiliation{Department of Engineering, University of Oxford}%

\date{\today}

\begin{abstract}
Machine learning has emerged as a significant approach to efficiently tackle electronic structure problems. Despite its potential, there is less guarantee for the model to generalize to unseen data that hinders its application in real-world scenarios. To address this issue, a technique has been proposed to estimate the accuracy of the predictions. This method integrates machine learning with self-consistent field methods to achieve both low validation cost and interpret-ability. This, in turn, enables exploration of the model's ability with active learning and instills confidence in its integration into real-world studies.

\end{abstract}

\maketitle

{\em Introduction:} 
Determining the electronic structure of a system is a central challenge in computational modeling. Despite quantum mechanics, the governing theory, having been discovered over a century ago, the exact solution to many-body electronic structure is still limited to only a handful of systems, even with the vast computational power available today. Density functional theory(DFT)\cite{kohnDensityFunctionalTheory1996,Kohn:1965zzb,Hohenberg:1964zz}, is a successful tool in computational chemistry that provides insights into molecules and materials at a reduced cost. Despite significant improvements in time complexity, the size of systems that can be effectively modeled using DFT is still limited due to the exponentially growing cost as the system size increases\cite{stephensInitioCalculationVibrational1994,zhaoM06SuiteDensity2008,perdewRationaleMixingExact1996}. As a result, it remains a challenge to scale DFT to practical and important systems with more than 1000 atoms. 

The integration of {\em machine learning} models has led to significant advancements in the field of electronic structure calculations for very large systems\cite{Carleo:2019ptp,damewoodRepresentationsMaterialsMachine2023,keithCombiningMachineLearning2021,unkeMachineLearningForce2021,Zhang:2023xgz,mahmoudLearningElectronicDensity2020}. The machine learning method builds surrogate models for traditional algorithms by directly learn from data \cite{lecunDeepLearning2015}. A successful example, ML potentials, which predict the atomic forces learned from DFT, make it possible to simulate millions of atoms with DFT-level accuracy \cite{zhangDeepPotentialMolecular2018,behlerGeneralizedNeuralNetworkRepresentation2007,lu86PFLOPSDeep2021}.

However, those methods are restricted to application in molecular dynamics and fail to serve as a general surrogates for DFT. 
Recently, learning the full solution to the DFT equations, has been  explored\cite{zhongTransferableEquivariantParameterization2022,Li:2021xnl,mahmoudLearningElectronicDensity2020,schuttUnifyingMachineLearning2019,gongPredictingChargeDensity2019,guDeePTBDeepLearningbased2023,yuQH9QuantumHamiltonian2023}. Predictions can be made towards the charge density or the effective Hamiltonian for the Kohn-Sham auxiliary system. These approaches are generally versatile and can yield various physical properties without the need for a self-consistency loop. In contrast to conventional methods which is very transferable, ML models are reliant on the data on which they were trained and may exhibit suboptimal performance when applied to target systems outside of their training data. Consequently, the precise estimation of prediction accuracy plays a critical role when employing ML models, particularly when exploring uncharted systems.

Uncertainty estimation \cite{zhuFastUncertaintyEstimates2023,imbalzanoUncertaintyEstimationMolecular2021,abdarReviewUncertaintyQuantification2021} is an active research topic in the field of machine learning. Existing methods estimate the distribution of the prediction with Bayesian Neural Networks or ensembles of model\cite{lakshminarayananSimpleScalablePredictive2017, pmlr-v48-gal16}. However, these methods have no guarantee of prediction accuracy and are typically used to produce confidence. We suggest that more physical constraint, rather than only statistical information, should be considered within the design of machine learning DFT surrogates to make it more reliable in unknown systems. 

In this letter, we introduce a full surrogate model to DFT that by design provides an estimation of the convergence accuracy of its prediction in a physics-informed manner. This is archived by following the concept of error vector defined in the direct inversion of the iterative subspace(DIIS) optimization method \cite{pulayConvergenceAccelerationIterative1980,pulayImprovedSCFConvergence1982} that examines the commutability of the one-electron reduced density matrix and the effective Hamiltonian matrix. In a standard DFT calculation, the DIIS error vector is minimized to get the solution to the electronic structure, thus also indicating the distance from the obtained trial solution to the accurate solution. We introduced the DIIS error into ML models and generalize it to the predicted matrices while avoiding the expensive self consistent calculation. Further, we constructed this criterion in a fully differentiable way, that can be used to train the model or calculate its gradient with respect to the atom coordinates, which enables an uncertainty driven active learning\cite{kulichenkoUncertaintydrivenDynamicsActive2023,imbalzanoUncertaintyEstimationMolecular2021}.

{\em Theory:} In a Kohn-Sham DFT calculation using the Linear Combination of Atomic Orbitals(LCAO), a set of localized basis functions is utilized to expand the single-particle wavefunction.\cite{sunPySCFPythonbasedSimulations2018,sunRecentDevelopmentsPySCF2020,liLargescaleInitioSimulations2016}. Under the selected basis, the observable operators can be represented by matrices:
\begin{equation}
    O_{ij} = \left\langle \chi_i \middle | \hat{O} \middle | \chi_j \right\rangle
\end{equation}
where $\{\chi_i\}$ is the set of the local basis. 

Following the DFT theories, one can always write the total energy of a system $E$ as a functional of electron density. The electron density, as an observable, can also be projected to the finite basis as the density matrix $D$ \cite{popleKohnShamDensityfunctional1992}, and minimizing the total energy yields the density matrix for ground state $\tilde D$:
\begin{equation}\label{eq:energy_functional}
   \tilde D = \text{argmin}_D E(D)
\end{equation}
To achieve the minima, one can solve the Kohn-Sham equation, which is a generalized eigenvalue problem \cite{ghojoghEigenvalueGeneralizedEigenvalue2023}:
\begin{equation} \label{equ-KS}
    HC = SCE
\end{equation}
where $H$ is the effective Hamiltonian matrix defined as the derivative of total energy with respect to the density matrix\cite{lehtolaOverviewSelfConsistentField2020}, $S$ is he overlap matrix originating from the non-orthogonality of the localized basis. $C$ and $E$ are the wavefunction coefficients and eigenenergies that can be solved from this equation if $H$ is converged to the ground state:
\begin{align}
    H(D) & \equiv \frac{\delta E}{\delta D}\label{def-hamiltonian}\\
    D(C) & \equiv C \Lambda C^T \equiv D(C(H))\label{def-density_matrix}
\end{align}
where eq. \ref{def-density_matrix} comes from the definition of electron density, and $\Lambda$ is the occupation matrix that is governed by the Pauli exclusion principle. If we find a pair of matrix $(\tilde D, \tilde H)$ that establish both equations at the same time, they are called converged and give the ground state of the target system.

Traditionally, the converged pair is found by solving Eq.(\ref{def-hamiltonian}) and Eq.(\ref{def-density_matrix}) repeatedly, until a fix point is found. This procedure is called the self consist field(SCF) iteration, which is usually computationally expensive. There have been many attempts to accelerate the procedure, and one of the most successful strategy is DIIS\cite{pulayConvergenceAccelerationIterative1980,pulayImprovedSCFConvergence1982}. In this strategy, besides a special interpolate algorithm that greatly improved the convergence, an error vector is defined as:
\begin{equation}\label{eq-diis_vec}
    e = HDS - SDH
\end{equation}
and the ground state is achieved when the error vector is minimized to be 0, thus it serves as an exact criterion of convergence as well.
In this case, a key relationship is that Eq.(\ref{eq-diis_vec}) would always be zero if all the matrices are generated from Eq.(\ref{equ-KS}) and Eq.(\ref{def-density_matrix}) without considering Eq.(\ref{def-hamiltonian}). So an extra implicit constrain is that $H$ in Eq.(\ref{eq-diis_vec}) must be constructed from $D$ in the same equation to include the full self-consistent conditions.

{\em DIIS for DFT surrogates:} 
The previous studies on electronic structure prediction have chosen the hamiltonian matrix\cite{schuttUnifyingMachineLearning2019,Li:2021xnl,zhongTransferableEquivariantParameterization2022} or the spatial electronic density\cite{jorgensenDeepDFTNeuralMessage2020} as their target, and take the atom positions as input. However, the constrain of Eq.(\ref{def-hamiltonian}) says that we cannot simply verify our predicted hamiltonian by diagnolizing it and construct a density matrix. Instead, having a predicted density matrix first and construct the consequent hamiltonian matrix would be valid. A recent study\cite{shaoMachineLearningElectronic2023} tried to predict the density matrix directly, but it didn't take the system configuration as input. It learns a mapping from the core potential matrix, which is also a valid form of DFT surrogate.

Considering all the constraints, there remains another problem: constructing Hamiltonian matrix from density matrix needs to evaluate the energy functional, which is just the most computational expensive step in SCF iterations. No matter how fast and cheap we got the electronic structure, we would have to do an $O(N^4)$ calculation to verify the result\cite{stephensInitioCalculationVibrational1994,zhaoM06SuiteDensity2008,perdewRationaleMixingExact1996}.. To avoid the expensive computation, we explore the approach where we predict both the Hamiltonian matrix and the density matrix using ML models. In addition, we adopt the hamiltonian formalism\cite{schuttUnifyingMachineLearning2019} that use the atom postitions as the input. Under this formalism, we can easily calculate the self-DIIS gradients with respect to atom positions since the model is fully differentiable. Due to the energy dimension of self-DIIS error, its gradients is related to a virtual atomic force that attempts to lower the uncertainly. By reversing the force, it becomes an active learning strategy naturally.

In practise, we define and construct the DIIS matrices from $H$ and $D$:
\begin{align}
    A = HDS \\
    B = SDH
\end{align}
and the model should minimize the difference between $A$ and $B$ to give accurate predictions. The way we define the difference can be chosen in different ways like cross-entropy and other common loss functions used in machine learning, not necessarily the element-wise MAE error. For a simple understanding, when the basis is orthognormal, the overlap matrix becomes an identity, thus minimizing the DIIS error indeed required the Hamiltonian matrix and the density matrix to have the same generalized eigenvectors.

\begin{figure}
    \centering
    \includegraphics[width=1\linewidth]{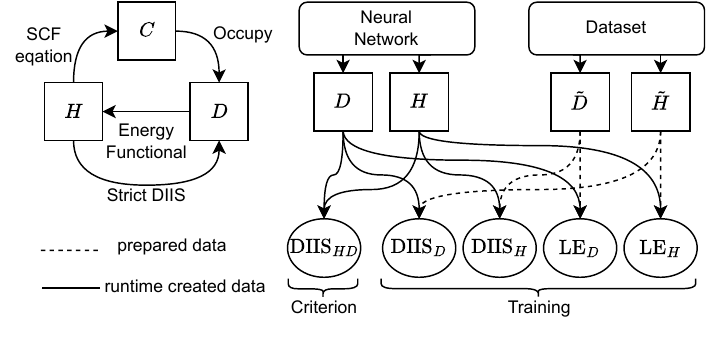}
    \caption{\textbf{The data stream under different methods.} Left:the SCF iteration and the construction of strict DIIS error vector. Right: The definition of different loss functions. Here DIIS means the error measures the difference between A and B, and LE is labeled error, that measures the difference between the predicted matrix and its label}
    \label{fig-workflow}
\end{figure}

As shown in Fig. \ref{fig-workflow}, we can define 3 different type of DIIS errors by using labeled matrices or predicted matrices to construct $A$ and $B$. If the labeled matrix and predicted matrix are used together, the DIIS error can be a regularization for the training loss function that helps the prediction to approach the convergence limit, and if both matrices are predicted ones, we call it the self-DIIS error, which just gives the convergence criterion for the model results. Notice that although we were discussed in the context of finite system, it is easy to generalize to extend systems by evaluating the self-DIIS error on each k-points, or even only on the gamma point in the brillouin zone\cite{simon2013oxford,martin2020electronic}.

Instead of using the hard constraint from Eq.(\ref{def-hamiltonian}), the approximation here is to use the soft constraint implied in training data. Although there are infinite pairs of $(D,H)$ which can lead to $A=B$, the probability of generating such a pair from random error is extremely small, especially when the model is likely to generate predictions in the neighbor of the exact solution with noisy error, as shown in the following experiments.

The ML model we used here is a Graph neural network(GNN), and is based on the message passing formalism\cite{gilmerNeuralMessagePassing2017} E(3) Symmetry is considered by tracing the group representation of the feature vectors in the model\cite{geigerE3nnEuclideanNeural2022}. The molecular structure is converted into a graph according the atom positions and their element, and we also utilize the atomic cluster expansion\cite{drautzAtomicClusterExpansion2019,dussonAtomicClusterExpansion2022,batatiaMACEHigherOrder2023} to capture the rich many-body information in local environments. The target matrices will be readout in the last layer and on each edge, and different readout heads are used for $H$ or $D$. Support for diversed basis sets is also supported with the multi-readout scheme. Our model is first neural surrogate to DFT matrices with multiple target and it achieved similar performance on hamiltonian matrix as previous works, and all details about our model is reported in the supplementary materials.


\begin{figure}
    \centering
    \includegraphics[width=1\linewidth]{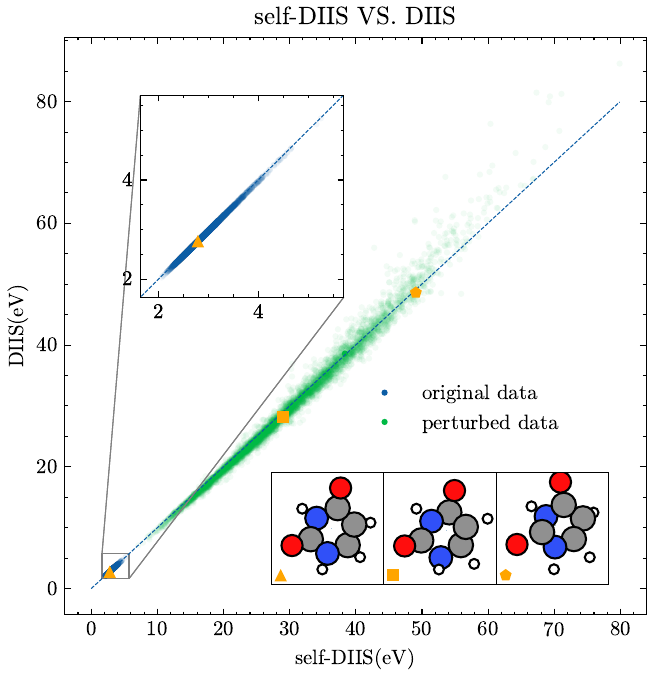}
    \caption{\textbf{The generalization ability of self-DIIS.} We evaluated the strict DIIS loss and the self-DIIS loss on both the original validation dataset and a purturbed dataset of the same molecular(uracil). The line is fitted only on the original dataset and the orange points represents the related data drawn at the bottom.}
    \label{fig-correlation}
\end{figure}

{\em The generalization ability of self-DIIS error:}
The approximation of the self-DIIS error with respect to the strict DIIS error become exact only in the limit of exact prediction. Hence the larger the error, the self-DIIS error is more likely to become inaccurate. To verify the robustness in extremely bad case that our criterion may give false positive results, we furthered explored our method on a out of distribution dataset. In the first experiment, we compare the label-free self-DIIS error with the labeled error on different datasets.


We used the RMD17\cite{christensenRoleGradientsMachine2020} and the QM9\cite{ruddigkeitEnumeration166Billion2012a,ramakrishnanQuantumChemistryStructures2014} dataset this experiment. The RMD17 dataset are trajectories of single molecules that are only different in atom position, while the QM9 contains different molecules that are different in both atom position and element components. There are 100,000 data for each RMD17 molecular and 133,675 molecules in QM9, and all labels are calculated and collected with PYSCF\cite{sunRecentDevelopmentsPySCF2020,sunPySCFPythonbasedSimulations2018}. We also noticed that a recent work published a new version of QM9 that take hamiltonian matrix into consideration\cite{yuQH9QuantumHamiltonian2023}, but didn't include density matrix with a small basis.

As shown in Fig. \ref{fig-correlation}, The predicted self-DIIS and the strict DIIS error fits very well in the trained region and it kept a very strong linear relationship in the region far from the original data distribution. The perturbed data is generated by adding random displacements to the original dataset, and larger displacements are not applied because it cause hard convergence when calculating labels with DFT. The largest DIIS error in the perturbed dataset is a magnitude larger than the origin dataset, where the self-DIIS is still very accurate. By analyzing the molecular structures in the datasets, the large self-DIIS error hints that the original dataset, simulated at room temperature, cannot describe a bond breaking case.

\begin{table}

\caption{\label{tab:correlation_results}\textbf{The linear regression correlation coefficient $R^2$ on different datasets.}  The first 2 columns are related to error distributions of strict convergence criterion conditioned on the self-DIIS error, and the last 2 columns report the error distribution of total energy and HOMO-LUMO gap conditioned on the strict DIIS error. The larger values between mean and std are highlighted, and we found that the correlation between self-DIIS and labeled error is dominated by their mean value, while the physical attributes are sensitive to the strict DIIS error in different way among different datasets.}

\begin{tabular}{lcc|cc||cc|cc}
\toprule
 & \multicolumn{2}{c}{DIIS} & \multicolumn{2}{c}{MAE} & \multicolumn{2}{c}{$E_{\text{tot}}$} & \multicolumn{2}{c}{$\Delta \epsilon$}  \\
 & mean & std & mean & std & mean & std & mean & std \\
\midrule
aspirin  & \textbf{0.99} & 0.12 & \textbf{0.99} & 0.61 & 0.52 & \textbf{0.65} & \textbf{0.26} & 0.22 \\
azobenz. & \textbf{0.99} & 0.08 & \textbf{0.97} & 0.64 & 0.68 & \textbf{0.68} & 0.37 & \textbf{0.58} \\
benzene  & \textbf{0.99} & 0.19 & \textbf{0.98} & 0.87 & 0.86 & \textbf{0.96} & 0.66 & \textbf{0.76} \\
ethanol  & \textbf{0.98} & 0.14 & \textbf{0.94} & 0.69 & \textbf{0.71} & 0.65 & \textbf{0.66} & 0.65 \\
malonal. & \textbf{0.99} & 0.13 & \textbf{0.94} & 0.76 & 0.36 & \textbf{0.55} & 0.55 & \textbf{0.75} \\
naphtha. & \textbf{0.99} & 0.09 & \textbf{0.98} & 0.51 & \textbf{0.82} & 0.53 & \textbf{0.68} & 0.62 \\
paracet. & \textbf{1.00} & 0.15 & \textbf{0.99} & 0.73 & 0.38 & \textbf{0.78} & \textbf{0.84} & 0.82 \\
salicyl. & \textbf{0.99} & 0.12 & \textbf{0.95} & 0.68 & 0.02 & \textbf{0.77} & 0.01 & \textbf{0.73} \\
toluene  & \textbf{0.99} & 0.06 & \textbf{0.95} & 0.45 & 0.50 & \textbf{0.71} & 0.29 & \textbf{0.18} \\
uracil   & \textbf{0.99} & 0.19 & \textbf{0.97} & 0.64 & \textbf{0.88} & 0.87 & 0.63 & \textbf{0.82} \\
QM9      & \textbf{1.00} & 0.36 & \textbf{0.94} & 0.84 & \textbf{0.89} & 0.87 & \textbf{0.88} & 0.84 \\
\bottomrule
\end{tabular}
\end{table}

{\em The correlation with physical attributes:}
The current machine learning models are not able to give predictions at the accuracy of a strict convergence, so the error distribution of the physical attributes at a loose convergence condition is critical in actual applications. Though the DIIS error is widely used as convergence criterion in traditional DFT calculation, it has never been used as a metric of error before, that its relationship to physical attributes is unknown. 

\begin{figure}
    \centering

    \includegraphics[width=1\linewidth]{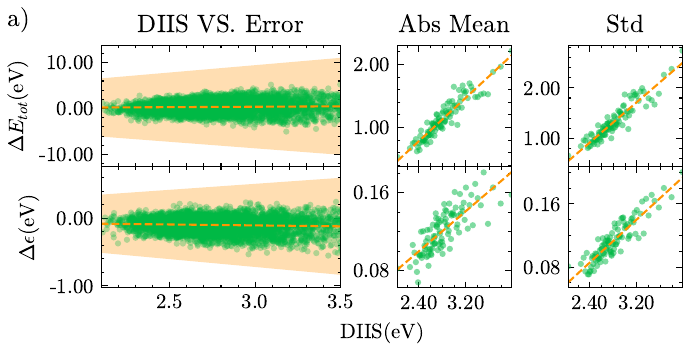}
    \includegraphics[width=1\linewidth]{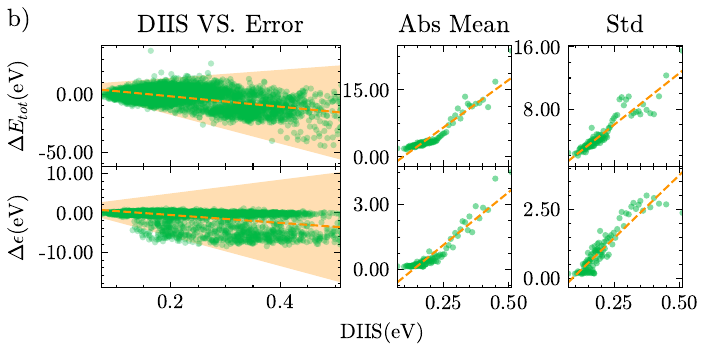}
    
    \caption{\textbf{The statistic results on uracil(a) and QM9(b).} The left panel showed the original data as green points, and the fitted mean value conditioned on DIIS error is drawn as the orange line. The shadow area is the 3$\sigma$ region estimated by the fitted standard deviation. The right 2 panels toke each group as a point, and the values are the related statistic result counted in that group. The orange line fitted directly on those points with a linear model, that provides with the statistics information used to draw the line and shadow on the left. On both datasets, that one is different in atom positions and another is different in element component, the error is well bound by the DIIS error in our test.}
    \label{fig-attributes_stat}
\end{figure}

In this experiment, we evaluate the error of strict DIIS, the total energy($E_\text{tot}$) and the HOMO-LUMO gap($\Delta \epsilon$) on the datasets, and group the data points according to their DIIS error. Then, the error distribution of $E_\text{tot}$ and $\Delta \epsilon$ are count in each group. Finally, a linear regression are applied to the mean value and the standard error of the groups respectively. Tow examples of such statistic procedure is illustrated in Fig. \ref{fig-attributes_stat}, and Tab. \ref{tab:correlation_results} shows the regression coefficients $R^2$ on all datasets. The correlation between the self-DIIS error and the labeled errors, including the physical strict DIIS error and the canonical element-wise mean absolute error(MAE) in machine learning, are count in the same way as well, and are shown in the first 2 columns of the table. The strong linear correlation hints that in most case, we can estimate the attribute error from the predicted self-DIIS error.

{\em Application in Molecular Dynamics:}
As a minimum example, we perform an ab-initio molecular dynamics simulation on the uracil dataset. The simulation is run with a predictor-corrector manner that the atomic force is calculated from the predicted density matrix if the self-DIIS error is lower than a threshold, otherwise an exact DFT calculation will be used. We run the simulation under a NVT ensemble with Berendsen temperature coupling
\cite{hunenbergerThermostatAlgorithmsMolecular2005}. The error threshold is set to 0.17 according to the count results shown in Fig. \ref{fig-correlation}, and the target $T$ is set to 1000K which is much higher than the temperature of the RMD17 dataset which is 300K. Some of the configurations in a high temperature are not likely to exist in a low temperature, it is expected that the surrogate model trained on a colder dataset will not have good performance in hot case.

\begin{figure}
    \centering
    \includegraphics[width=1\linewidth]{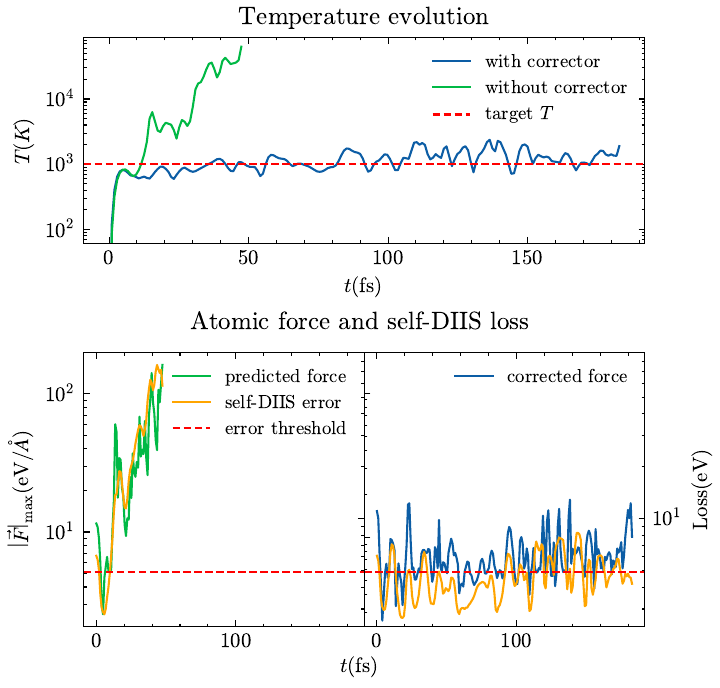}
    \caption{\textbf{Running AIMD with and without error conditioned corrector.} A simulation with our predictor-corrector strategy and another simulation use the model prediction directly is drawn together. The left panel, the corrected one relaxed to the target temperature successfully while the other one diverged. The right panel shows the self-DIIS error and the max force applied to the atoms in each time step. In the failed case, the predicted force become exotic large with the increase of self-DIIS loss. While in the corrected case, the force is calculated from DFT results once the self-DIIS loss is above the threshold line, and the force was kept in a reasonable range.}
    \label{fig:md_compare}
\end{figure}

Fig. \ref{fig:md_compare} showed the simulation result. The simulation without DFT correction failed  because the surrogate model gave bad prediction in some case, and the predictor-corrector simulation kept fluctuating around the target $T$. The fluctuation of $T$ in this single molecular case is mainly related to the oscillation of the chemical bounds in the molecule. The distance between atoms changes periodically, that caused the periodic change of self-DIIS error, so the self-DIIS error fluctuates synchronously with the temperature. It proves that the criterion successfully kept the simulation stable by filtering out bad predictions, which makes the ML model applicable in an uncovered scene, and further pointed out the data necessary to be considered in such case.

{\em Discussion:} Applying machine learning technology to scientific problems has gathered significant attention in the past few years, while it remains not generally usable in actually studies, mainly due to their nature of being black boxes. Our work aims at elevate them as useful tools in studies by giving a meaningful while cheap accuracy criterion. The core idea here is to take advantage of both data-driven approaches, which is faster, and physics driven approaches, which is explainable. We expect that similar improvements would be found beside the electronic structure problems we discussed in this study. Further, exploring more usage cases of the current study, like the mentioned uncertainty-driven active learning is very interesting as well.


\begin{acknowledgments}
The authors thank Mao Su, Yuqiang Li and Shengdu Chai for helpful discussions. This study is supported by Shanghai Artificial Intelligence Laboratory. All the datasets and code will be published soon after acceptance.
\end{acknowledgments}

\appendix

\nocite{*}

\bibliography{pcmds1}

\end{document}